%% file: naaclhlt2019.tex
\documentclass[11pt,a4paper]{article}
\usepackage[hyperref]{naaclhlt2019}
\usepackage{times}
\usepackage{latexsym}

\usepackage{url}

\aclfinalcopy 

\usepackage{enumitem}
\usepackage{amsmath}
\usepackage{booktabs}
\usepackage{graphicx}

\usepackage{resizegather}

\DeclareMathOperator*{\argmin}{argmin}

\title{Discontinuous Constituency Parsing\\
  with a Stack-Free Transition System and a Dynamic Oracle}

\author{Maximin Coavoux\Thanks{ Work done at the University of Edinburgh.}\\
    Naver Labs Europe\\
    {\tt maximin.coavoux@naverlabs.com} \\\And
    Shay B. Cohen \\
    ILCC, School of Informatics \\
    University of Edinburgh \\
    {\tt scohen@inf.ed.ac.uk} \\}

\date{}

\begin{document}

\input{article.tex}

\interlinepenalty=1000

\bibliography{naaclhlt2019}
\bibliographystyle{acl_natbib}

\input{appendix.tex}

\end{document}

%% file: article.tex
\maketitle
\begin{abstract}
    We introduce a novel transition system for
    discontinuous constituency parsing.
    Instead of storing
    subtrees in a stack --i.e.\ a data
    structure with linear-time \textit{sequential access}--
    the proposed system uses a \textit{set} 
    of parsing items, with constant-time \textit{random access}.
    This change makes it possible to construct any discontinuous
    constituency tree in exactly $4n-2$ transitions
    for a sentence of length $n$, whereas existing
    systems need a quadratic number of transitions to derive some structures.
    At each parsing step, the parser considers every item
    in the set to be combined with a \textit{focus} item
    and to construct a new constituent in a bottom-up fashion.
    The parsing strategy is based on the assumption that most
    syntactic structures can be parsed incrementally
    and that the set --the memory of the parser-- remains
    reasonably small on average.
    Moreover, we introduce a dynamic
    oracle for the new transition system, and present the
    first experiments in discontinuous constituency
    parsing using a dynamic oracle.
    Our parser obtains state-of-the-art results on three
    English and German discontinuous treebanks.
\end{abstract}

\section{Introduction}

\begin{figure}[t]
    \resizebox{\columnwidth}{!}{
        \includegraphics{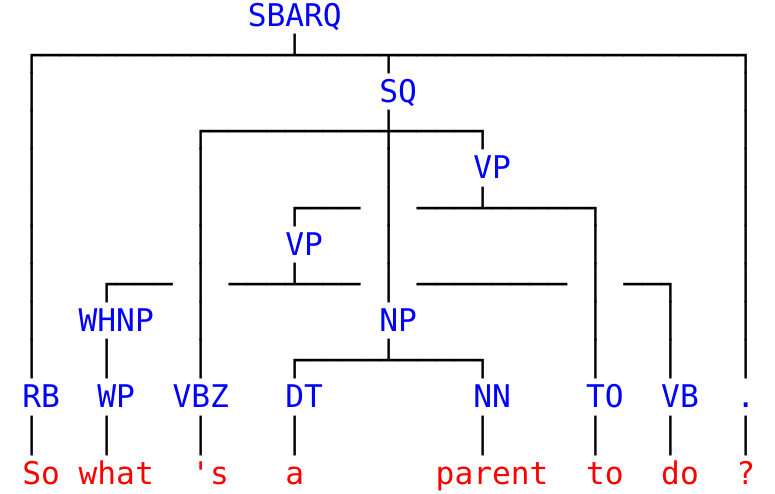}
    }
    \caption{Discontinuous constituency tree
      from the Discontinuous Penn treebank.}
      \label{fig:disco-tree}
\end{figure}

Discontinuous constituency trees extend standard
constituency trees by allowing crossing branches
to represent long distance dependencies,
such as the \textit{wh}-extraction in Figure~\ref{fig:disco-tree}.
Discontinuous constituency trees can be seen
as derivations of Linear Context-Free Rewriting Systems \cite[LCFRS,][]{P87-1015}, a class of formal
grammars more expressive than context-free grammars, which makes them much
harder to parse.
In particular, exact CKY-style LCFRS parsing has an $\mathcal{O}(n^{3f})$
time complexity where $f$ is the fan-out
of the grammar \cite{kallmeyer-book}.

A natural alternative to grammar-based chart parsing is transition-based parsing,
that usually relies on fast approximate decoding methods such as greedy search or beam search.
Transition-based discontinuous parsers construct discontinuous constituents
by reordering terminals with the \textsc{swap} action \cite{versley:2014:SPMRL-SANCL,DBLP:journals/corr/Versley14,maier:2015:ACL-IJCNLP,maier-lichte:2016:DiscoNLP,stanojevic-garridoalhama:2017:EMNLP2017},
or by using a split stack and the \textsc{gap} action
to combine two non-adjacent constituents \cite{coavoux-crabbe:2017:EACLlong,1902.08912}.
These proposals represent the memory of the parser (i.e.\ the tree
fragments being constructed) with data structures with
\textbf{linear-time sequential access} (either a stack, or
a stack coupled with a double-ended queue).
As a result, these systems need to perform 
at least $n$ actions to construct a new constituent
from two subtrees separated by $n$ intervening subtrees.
Our proposal aims at avoiding this cost when constructing discontinuous constituents.

\begin{figure}[t]
    \resizebox{\columnwidth}{!}{
    \includegraphics{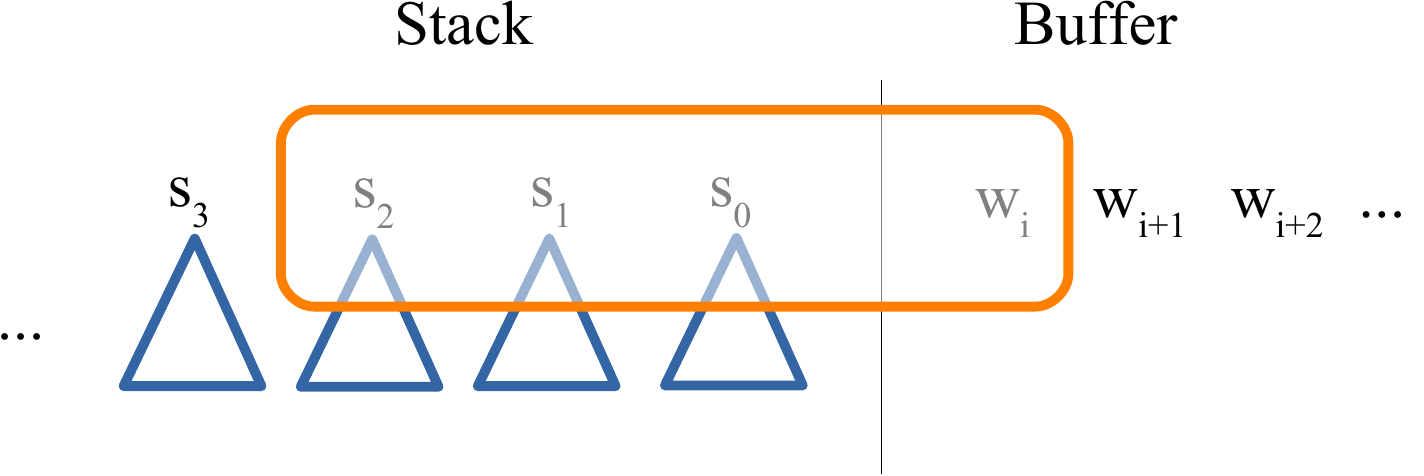}}
\noindent\rule{\columnwidth}{0.5pt}
\vspace{0.5em}

    \resizebox{\columnwidth}{!}{
    \includegraphics{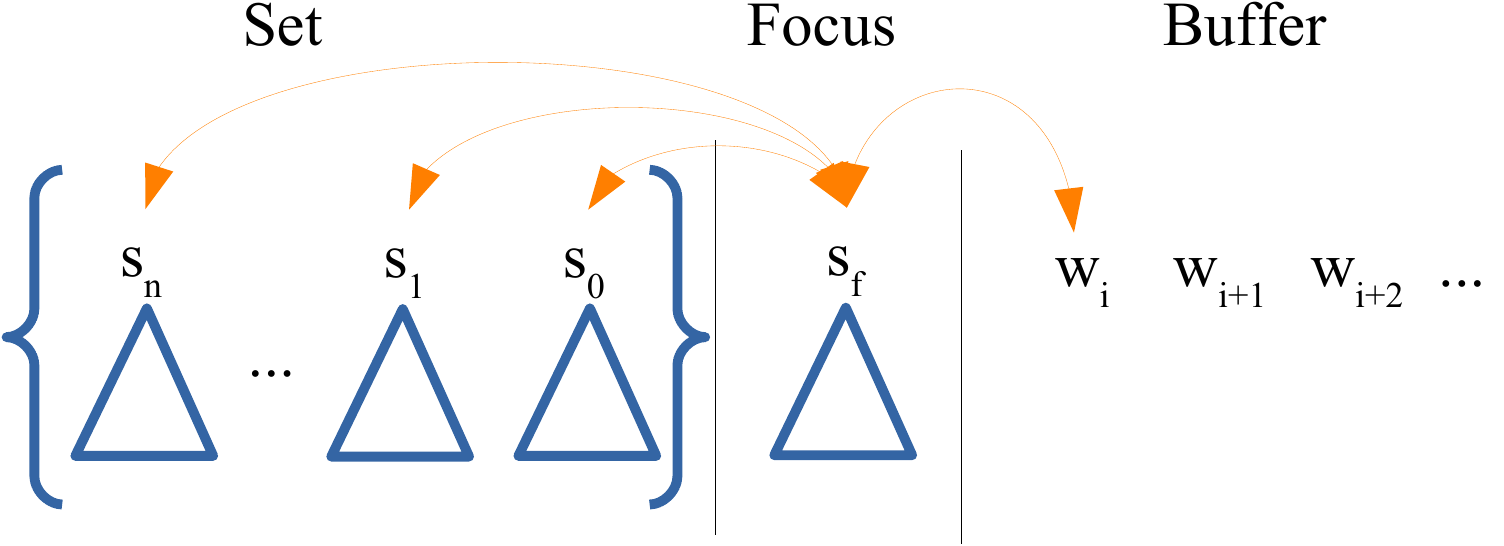}}
\caption{In a stack-based system like shift-reduce-swap (upper part),
the parser extracts features from a \textit{local}
region of the configuration (orange part),
to predict the next action such as: construct a new tree
with label X and children $s_0$ and $s_1$ (\textsc{reduce-X}).
In our proposed set-based system (lower part), we consider every item
in the set to be combined bottom-up with the focus item $s_f$
and score independently each possible transition.}
\label{fig:intuition}
\end{figure}

\newcommand{\spa}{\;\;}

\begin{table*}
\begin{center}
\resizebox{\textwidth}{!}{
\begin{tabular}{lrll}
    \toprule
    Initial configuration   & \multicolumn{3}{l}{$(\emptyset, \spa \texttt{null}, \spa 0, \spa \emptyset):0$} \\
    Goal configuration      & \multicolumn{3}{l}{$(\emptyset, \spa \{0,1,\dots,n-1\}, \spa n, \spa C):4n-2$} \\
    \midrule
    Structural actions      & Input & Output & Precondition\\
    \midrule
    \textsc{shift}          & $(S, \spa s_f, \spa i, \spa C):j$
                            & $\Rightarrow (S \cup \{s_f\}, \spa \{i\}, \spa i+1, \spa C):j+1$ 
                            & $i < n$, $j$ is even\\
    \textsc{combine}-$s$    & $(S, \spa s_f, \spa i, \spa C):j$
                            & $\Rightarrow (S - s, \spa s_f \cup s, \spa i, \spa C):j+1$ 
                            & $s \in S$, $j$ is even\\
    \midrule
    Labelling actions & & & \\
    \midrule
    \textsc{label-X}        & $(S, \spa s_f, \spa i, \spa C):j$ 
                            & $\Rightarrow (S, \spa s_f, \spa i, \spa C \cup \{(X, \spa s_f)\}):j+1$ 
                            & $j$ is odd\\
    \textsc{no-label}       & $(S, \spa s_f, \spa i, \spa C):j$
                            & $\Rightarrow (S, \spa s_f, \spa i, \spa C):j+1$ 
                            & $i \neq n$ or $S \neq \emptyset$, $j$ is odd\\
    \bottomrule
\end{tabular}
}
\end{center}
\caption{Set-based transition system description. Variable $j$ is the number of steps performed
since the start of the derivation.}
\label{tab:transition-system}
\end{table*}

We design a novel transition system in which
a discontinuous constituent is constructed in a single step,
without the use of reordering actions such as \textsc{swap}.
The main innovation is that
the memory of the parser is not represented
by a stack, as is usual in shift-reduce systems,
but by an unordered \textbf{random-access} set.
The parser considers every constituent in the current
memory to construct a new constituent in a bottom-up fashion,
and thus instantly models interactions between parsing items
that are not adjacent.
As such, we describe a left-to-right parsing model that deviates from the standard stack-buffer setting,
a legacy from pushdown automata and classical parsing algorithms for context-free grammars.

Our contributions are summarized as follows:
\begin{itemize}[noitemsep]
    \item We design a novel transition system
    for discontinuous constituency parsing,
    based on a memory represented by a \textit{set} of items,
    and that derives any tree in exactly $4n-2$ steps for a sentence of length $n$;
    \item  we introduce the first dynamic oracle for discontinuous constituency parsing;
    \item  we present an empirical evaluation of the transition system
    and dynamic oracle on two German and one English discontinuous treebanks.
\end{itemize}
The code of our parser is released as an open-source project at \url{https://gitlab.com/mcoavoux/discoparset}.

\newcommand{\ra}{$\Rightarrow$}
\begin{table*}
    \resizebox{\textwidth}{!}{
    \begin{tabular}{rrcll}
    \toprule
    Even action & Set ($S$) & Focus ($s_f$) & Buffer & Odd action \\
    \midrule
                                           &\{\} & \texttt{none} & So what 's a parent to do ? & \\
\ra \textsc{sh}\ra    &\{\} & \{So\} & what 's a parent to do ?  & \ra \textsc{no-label} \\
\ra \textsc{sh}\ra   &\{\{So\}$_0$\} &  \{what\} & 's a parent to do ? & \ra \textsc{label-WHNP} \\
\ra \textsc{sh}\ra    &\{\{So\}$_0$,  \{what\}$_1$\} &  \{'s\} & a parent to do ? & \ra \textsc{no-label} \\
\ra \textsc{sh}\ra    &\{\{So\}$_0$,  \{what\}$_1$, \{'s\}$_2$\} & \{a\} & parent to do ? & \ra \textsc{no-label} \\
\ra \textsc{sh}\ra    &\{\{So\}$_0$,  \{what\}$_1$, \{'s\}$_2$, \{a\}$_3$\} & \{parent\} & to do ? & \ra \textsc{no-label} \\
\ra \textsc{comb-3}\ra    &\{\{So\}$_0$, \{what\}$_1$, \{'s\}$_2$\} & \{a parent\} & to do ? & \ra \textsc{label-NP} \\
\ra \textsc{comb-2}\ra     &\{\{So\}$_0$, \{what\}$_1$\} & \{'s a parent\} & to do ? & \ra \textsc{no-label} \\
\ra \textsc{sh}\ra    &\{\{So\}$_0$,  \{what\}$_1$, \{'s a parent\}$_2$\}& \{to\} & do ? & \ra \textsc{no-label} \\
\ra \textsc{sh}\ra     &\{\{So\}$_0$,  \{what\}$_1$, \{'s a parent\}$_2$, \{to\}$_5$\}& \{do\} & ? & \ra \textsc{no-label} \\
\ra \textsc{comb-1}\ra     &\{\{So\}$_0$, \{'s a parent\}$_2$, \{to\}$_5$\}& \{what do\} & ? & \ra \textsc{label-VP} \\
\ra \textsc{comb-5}\ra    &\{\{So\}$_0$, \{'s a parent\}$_2$\}& \{what to do\} & ? &  \ra \textsc{label-VP} \\
\ra \textsc{comb-2}\ra    &\{\{So\}$_0$\}& \{what 's a parent to do\} & ? & \ra \textsc{label-SQ} \\
\ra \textsc{comb-0}\ra    &\{\}& \{so what 's a parent to do\} & ? &  \ra \textsc{no-label} \\
\ra \textsc{sh}\ra     &\{\{So what 's a parent to do\}$_0$\} & \{?\} & & \ra \textsc{no-label}\\
\ra \textsc{comb-0}\ra & \{\} & \{So what 's a parent to do ?\} & & \ra \textsc{label-SBARQ} \\
    \bottomrule
    \end{tabular}}
\caption{Full derivation for the sentence in Figure~\ref{fig:disco-tree}.
As a convention, we index elements in the set with their left-index
and use \textsc{comb}-$i$ to denote \textsc{comb}-$s_i$.
We also use tokens instead of their indexes for better legibility.}
\label{tab:derivation}
\end{table*}

\section{Set-based Transition System}
\label{sec:system}

\paragraph{System overview}
We propose to represent the memory of the parser by (i) a set of parsing items
and (ii) a single \textit{focus item}.
Figure~\ref{fig:intuition} (lower part) illustrates a configuration in our system.
The parser constructs a tree with two main actions:
shift the next token to make it the new focus item (\textsc{shift}),
or combine any item in the set with the focus item to make
a new constituent bottom-up (\textsc{combine} action).

Since the memory is not an ordered data structure, the parser
considers equally every pending parsing item, and thus
constructs a discontinuous constituent in a single step,
thereby making it able to construct any discontinuous tree
in $\mathcal{O}(n)$ transitions.

The use of an unordered random-access data structure to represent
the memory of the parser also leads to a major change
for the scoring system (Figure~\ref{fig:intuition}).
Stack-based systems use a \textit{local view} of a parsing configuration
to extract features and score actions: features only rely on the few topmost
elements on the stack and buffer.
The score of each transition depends on the totality of this local view.
In constrast, we consider equally every item in the set,
and therefore rely on a \textit{global view}
of the memory (Section~\ref{sec:net}).
However, we score each possible combinations independently:
the score of a single combination only depends on the two constituents
that are combined, regardless of the rest of the set.

\subsection{System Description}

\paragraph{Definitions}

We first define an \textit{instantiated (discontinuous)
constituent} $(X, s)$ as a nonterminal label $X$ associated
with a set of token indexes $s$.
We call $\min(s)$ the \textit{left-index} of $c$ and $\max(s)$ its \textit{right-index}.
For example in Figure~\ref{fig:disco-tree}, the two VPs are
respectively represented by (VP, \{1, 6\}) and (VP, \{1, 5, 6\}),
and they have the same right index (6) and left index (1). 

A \textit{parsing configuration} is a quadruple $(S, s_f, i, C)$ where:
\begin{itemize}[noitemsep]
    \item $S$ is a set of sets of indexes and represents the memory of the parser;
    \item $s_f$ is a set of indexes called the \textit{focus item}, and satisfies $\max(s_f)=i-1$;
    \item $i$ is the index of the next token in the buffer;
    \item $C$ is a set of instantiated constituents.
\end{itemize}
Each new constituent is constructed bottom-up from the \textit{focus}
item and another item in the set $S$.

\paragraph{Transition set}
Our proposed transition system is based on the following
types of actions:
\begin{itemize}[noitemsep]
    \item \textsc{shift} constructs a singleton containing
    the next token in the buffer and assigns it as the new focus item.
    The former focus item is added to $S$.
    \item \textsc{combine}-$s$ computes the union
    between the focus item $s_f$ and another
    item $s$ from the set $S$, to form the new focus item $s \cup s_f$.
    \item \textsc{label-X} instantiates a new constituent $(X, s_f)$
    whose yield is the set of indexes in the focus item $s_f$.
    \item \textsc{no-label} has no effect; its semantics is that the current
    focus set is not a constituent.
\end{itemize}
Following \newcite{cross-huang:2016:EMNLP2016},
transitions are divided into \textbf{structural} actions (\textsc{shift}, \textsc{combine}-$s$)
and \textbf{labelling} actions (\textsc{label-X}, \textsc{no-label}).
The parser may only perform a structural action on an even step
and a labelling action on an odd step.
For our system, this distinction has the crucial advantage of keeping
the number of possible actions low at each parsing step,
compared to a system that would perform a \textsc{combine} action and a labelling
action in a single \textsc{reduce}-$s$-X action.\footnote{In such a case, we would need to score $|S|\times|N| +1$ actions, where $N$ is the set of nonterminals, instead of $|S|+1$ actions for our system.}

Table~\ref{tab:transition-system} presents each action
as a deduction rule associated with preconditions.
In Table~\ref{tab:derivation}, we describe how to derive
the tree from Figure~\ref{fig:disco-tree}.

\subsection{Oracles}

Training a transition-based parser requires an oracle,
i.e.\ a function that determines what the best action
is in a specific parsing configuration to serve as
a training signal.
We first describe a static oracle that provides a canonical derivation
for a given gold tree.
We then introduce a dynamic oracle that determines what the best action
is in any parsing configuration.

\subsubsection{Static Oracle}

Our transition system exhibits a fair amount of \textit{spurious ambiguity},
the ambiguity exhibited by the existence of many possible derivations for a single tree.
Indeed, since we use an unordered memory, an $n$-ary constituent (and more generally a tree)
can be constructed by many different transition sequences.
For example, the set \{0, 1, 2\} might be constructed by
combining 
\begin{itemize}[noitemsep]
    \item \{0\} and \{1\} first, and the result with \{2\}; or
    \item \{1\} and \{2\} first, and the result with \{0\}; or
    \item \{0\} and \{2\} first, and the result with \{1\}.
\end{itemize}

Following \citet{DBLP:journals/corr/abs-1206-6735}, we eliminate
spurious ambiguity by selecting a canonical derivation for a gold tree.
In particular, we design the static oracle 
(i) to apply \textsc{combine} as soon as possible in order to minimize
the size of the memory (ii) to combine preferably with the most
recent set in the memory when several combinations are possible.
The first choice is motivated by properties of our system:
when the memory is smaller, there are fewer choices, therefore decisions
are simpler and less expensive to score.

\subsubsection{Dynamic Oracle}
\label{sec:dyno}

Parsers are usually trained to predict the gold sequence of actions, using a static oracle.
The limitation of this method is that the parser only sees
a tiny portion of the search space at train time and only trains
on gold input (i.e.\ configurations obtained after performing gold actions).
At test time, it is in a different situation due to error propagation:
it must predict what the best actions are in configurations
from which the gold tree is probably no longer reachable.

To alleviate this limitation, \newcite{goldberg-nivre:2012:PAPERS}
proposed to train a parser with a \textit{dynamic oracle}, an oracle 
that is defined for any parsing configuration
and outputs the set of best actions to perform.
In contrast, a \textit{static oracle} is deterministic
and is only defined for gold configurations.

Dynamic oracles were proposed for a wide range of dependency parsing transition systems
\citep{Q13-1033,gomezrodriguez-sartorio-satta:2014:EMNLP2014,gomezrodriguez-fernandezgonzalez:2015:ACL-IJCNLP},
and later adapted to constituency parsing
\cite{coavoux-crabbe:2016:P16-1,cross-huang:2016:EMNLP2016,DBLP:journals/corr/abs-1804-07961,fernndezgonzlez-gmezrodrguez:2018:EMNLP}.

In the remainder of this section, we introduce a dynamic oracle
for our proposed transition system.
It can be seen as an extension of the oracle of
\newcite{cross-huang:2016:EMNLP2016}
to the case of discontinuous parsing.

\paragraph{Preliminary definitions}
For a parsing configuration $c$, the relation $c \vdash c'$ 
holds iff $c'$ can be derived from $c$ by a single transition.
We note $\vdash^*$ the reflexive and transitive closure of $\vdash$.
An instantiated constituent $(X, s)$ is \textbf{reachable}
from a configuration $c=(S, s_f, i, C)$ iff there
exists $c'=(S', s'_f, i', C')$ such that $(X, s)\in C'$ and $c \vdash^* c'$.
Similarly, a set of constituents $t$ (possibly a full discontinuous constituency tree)
is reachable iff there exists a configuration $c'=(S', s'_f, i', C')$
such that $t \subseteq C'$ and $c \vdash^* c'$.
We note $\mathsf{reach}(c, t^*)$ the set of constituents
that are (i) in the gold set of constituents~$t^*$
(ii) reachable from $c$. 

We define a total order $\preceq$ on index sets:
\begin{equation*}
    s \preceq s' \Leftrightarrow \left\{
    \begin{array}{l}
    \max(s) < \max(s'), \\
    \textbf{or} \\
    \max(s) = \max(s') \\\text{and } s \subseteq s'.
    \end{array} \right.
\end{equation*}
This order naturally extends to the constituents of a tree:
$(X, s) \preceq (X', s')$ iff $s \preceq s'$.
If $(X, s)$ precedes $(X', s')$, then $(X, s)$
must be constructed before $(X', s')$.
Indeed, since the right-index of the focus item is non-decreasing
during a derivation (as per the transition definitions),
constituents are constructed in the order
of their right-index (first condition).
Moreover, since the algorithm is bottom-up, a constituent must
be constructed before its parent (second condition).

From a configuration $c=(S, s_f, i, C)$ at an odd step,
a constituent $(X, s_g) \notin C$ is reachable 
iff both the following properties hold:
\begin{enumerate}[noitemsep]
    \item $\max(s_f) \leq \max(s_g)$; \label{property1}
    \item $\forall s \in S\cup \{ s_f \}, (s \subseteq s_g) \text{ or } (s \cap s_g = \emptyset)$. \label{property2}
\end{enumerate}
Condition~\ref{property1} is necessary because the parser can only construct
new constituents $(X, s)$ such that $s_f \preceq s$.
Condition~\ref{property2} makes sure that
$s_g$ can be constructed from a union of elements from $S \cup \{s_f\}$,
potentially augmented with terminals from the bufffer: $\{i, i+1, \dots, \max(s_g)\}$.

Following \newcite{cross-huang:2016:EMNLP2016}, we define $\mathsf{next}(c, t^*)$
as the smallest reachable gold constituent from a configuration $c$.
Formally:
\begin{equation*}
    \mathsf{next}(c, t^*) = \argmin_{\preceq} \mathsf{reach}(c, t^*).
\end{equation*}

\paragraph{Oracle algorithm}
We first define the oracle $\mathsf{o}$ for the odd step
of a configuration $c = (S, s_f, i, C)$:
\begin{equation*}
    \mathsf{o_{odd}}(c, t^*) =
    \left\{\begin{array}{ll}
    \textsc{\{label-X\}}& \text{if } \exists (X, s_f) \in t^*,\\
    \textsc{\{no-label\}} & \text{otherwise.}
    \end{array}
    \right.
\end{equation*}

For even steps, assuming $\mathsf{next}(c, t^*)= (X, s_g)$,
we define the oracle as follows:
\begin{gather*}
    \mathsf{o_{even}}(c, t^*) =
    \left\{\begin{array}{r}
    \multicolumn{1}{l}{\{\textsc{comb-$s$} | (s_f \cup s) \subseteq s_g\}} \\
    \text{if } \max(s_g) = \max(s_f), \\
    \{\textsc{comb-$s$} | (s_f \cup s) \subseteq s_g\} \cup  \{\textsc{sh}\} \\ 
    \text{if } \max(s_g) > \max(s_f).
    \end{array}
    \right.
\end{gather*}
We provide a proof of the correctness of the oracle in Appendix~\ref{sec:correctness}.

\section{A Neural Network based on Constituent Boundaries}
\label{sec:net}

We first present an encoder that computes
context-aware representations of tokens
(Section~\ref{sec:token}).
We then discuss how to compute
the representation of a set of tokens (Section~\ref{sec:constituent}).
We describe the action scorer (Section~\ref{sec:actions}),
the POS tagging component (Section~\ref{sec:tagger}),
and the objective function (Section~\ref{sec:objective}).

\subsection{Token Representations}
\label{sec:token}

As in recent proposals in dependency and constituency parsing
\cite{cross-huang:2016:P16-2,TACL885},
our scoring system is based on a
sentence transducer that constructs a context-aware
representation for each token.

Given a sequence of tokens $x_1^n = (x_1, \dots, x_n)$,
we first run a single-layer character bi-LSTM encoder $\mathbf c$ to obtain
a character-aware embedding $\mathbf c(x_i)$ for each token.
We represent a token $x_i$ as the concatenation
of a standard word embedding $\mathbf e(x_i)$
and the character-aware embedding:
$\mathbf w_{x_i}=[\mathbf c(x_i);\mathbf e(x_i)].$

Then, we run a 2-layer bi-LSTM transducer over the sequence of token representations:
\begin{align*}
    (\mathbf h_1^{(1)}, \dots, \mathbf h_n^{(1)}) &= \text{bi-LSTM}(\mathbf w_{x_1}, \dots, \mathbf w_{x_n}), \\
    (\mathbf h_1^{(2)}, \dots, \mathbf h_n^{(2)}) &= \text{bi-LSTM}(\mathbf h_1^{(1)}, \dots, \mathbf h_n^{(1)}).
\end{align*}
The parser uses the context-aware token representations $\mathbf h_i^{(2)}$
to construct vector representations of sets or constituents.

\subsection{Set Representations}
\label{sec:constituent}

An open issue in neural discontinuous parsing is the representation
of discontinuous constituents.
In projective constituency parsing, it has become standard
to use the boundaries of constituents
\cite{hall-durrett-klein:2014:P14-1,crabbe:2015:EMNLP,durrett-klein:2015:ACL-IJCNLP},
an approach that proved very successful with bi-LSTM token representations
\cite{cross-huang:2016:EMNLP2016,stern-andreas-klein:2017:Long}.

Although constituent boundary features improves discontinuous
parsing \cite{coavoux-crabbe:2017:EACLlong}, relying only on the left-index and the right-index
of a constituent has the limitation of ignoring gaps inside a constituent.
For example, since the two VPs in Figure~\ref{fig:disco-tree}
have the same right-index and left-index, they would
have the same representations.
It may also happen that constituents with identical
right-index and left-index do not have the same labels.

We represent a (possibly partial) constituent with the
yield $s$, by computing~4 indexes from~$s$: $(\min(s), \max(s), \min(\overline{s}), \max(\overline{s}))$.
The set $\overline{s}$ represents the gap in $s$, i.e.\ the
tokens between $\min(s)$ and $\max(s)$ that are not in the yield of $s$:
\begin{equation*}
    \overline{s} = \{ i | \min(s) < i < \max(s) \text{ and } i \notin s \}.
\end{equation*}
Finally, we extract the corresponding token representations of
the~4 indexes and concatenate them to form the vector representation $\mathbf r(s)$ of $s$:
\begin{equation*}
    \mathbf r(s) = [\mathbf h^{(2)}_{\min(s)}; \mathbf h^{(2)}_{\max(s)};
                    \mathbf h^{(2)}_{\min(\overline{s})}; \mathbf h^{(2)}_{\max(\overline{s})}].
\end{equation*}
For an index set that does not contain a gap,
we have $\overline{s} = \emptyset$.
To handle this case, we use a parameter vector $\mathbf h_{\mathsf{nil}}$,
randomly initialized and learned jointly with the network,
to embed $\max(\emptyset) = \min(\emptyset) = \mathsf{nil}$.

For example, the constituents (VP, \{1, 6\}) and (VP, \{1, 5, 6\}) will be respectively vectorized as:
\begin{align*}
\mathbf r(\{1, 6\}) &= [\mathbf h^{(2)}_{1}; \mathbf h^{(2)}_{6}; \mathbf h^{(2)}_{2}; \mathbf h^{(2)}_{5}], \\
\mathbf r(\{1, 5, 6\}) &= [\mathbf h^{(2)}_{1}; \mathbf h^{(2)}_{6}; \mathbf h^{(2)}_{2}; \mathbf h^{(2)}_{4}].
\end{align*}
This representation method makes sure that two distinct index sets have distinct representations,
as long as they have at most one gap each.
This property no longer holds if one index sets has more than one gap.

\subsection{Action Scorer}
\label{sec:actions}

For each type of action --structural or labelling--
we use a feedforward network with two hidden layers.

\paragraph{Structural actions}
At structural steps, for a configuration $c=(S, s_f, i, C)$,
we need to compute the score of $|S|$ \textsc{combine}
actions and possibly a \textsc{shift} action.
In our approach, the score of a combine-$s$ action
only depends on $s$ and $s_f$ and is independent of the rest
of the configuration (i.e.\ other items in the set).
We first construct input matrix $M$ as follows:
\begin{align*}
    M &=    \left( 
            \begin{array}{cccc}
                \mathbf r(s_1) & \cdots & \mathbf r(s_n) & \mathbf r(\{i\}) \\
                \mathbf r(s_f) & \cdots & \mathbf r(s_f) & \mathbf r(s_f)  \\
            \end{array}
            \right).
\end{align*}
Each of the first $n$ columns of matrix $M$ represents the input for 
a \textsc{combine} action, whereas the last column is the input for the \textsc{shift} action.
We then compute the score of each structural action:
\begin{align*}
    P(\cdot | c) &= \text{Softmax}(\text{FF}_{s}(M)),
\end{align*}
where $\text{FF}_{s}$ is a feedforward network with two hidden layers,
a $\tanh$ activation and a single output unit.
In other words, it outputs a single scalar for each column vector of matrix $M$.
This part of the network can be seen as an attention mechanism, where the focus
item is the query, and the context is formed by the items in the set and the first element in the buffer.

\paragraph{Labelling actions}
We compute the probabilities of labelling actions as follows:
\begin{align*}
    P(\cdot |s_f)           &= \text{Softmax}(\text{FF}_l(\mathbf r(s_f))),
\end{align*}
where $\text{FF}_l$ is a feedforward network with two hidden layers activated with the $\tanh$ function,
and $|N|+1$ output units, where $N$ is the set of nonterminals.

\subsection{POS Tagger}
\label{sec:tagger}

Following \newcite{coavoux-crabbe:2017:EACLshort}, we use the first
layer of the bi-LSTM transducer as input to a Part-of-Speech (POS) tagger that is learned
jointly with the parser.
For a sentence $x_1^n$, we compute the probability of a sequence of POS tags $t_1^n=(t_1, \dots, t_n)$ as follows:
\begin{equation*}
    P(t_1^n|x_1^n) = \prod_{i=1}^n \text{Softmax}(\mathbf W^{(t)} \cdot \mathbf h^{(1)}_i + \mathbf b^{(t)})_{t_i},
\end{equation*}
where $\mathbf W^{(t)}$ and $\mathbf b^{(t)}$ are parameters.

\begin{table*}[t]
    \begin{center}
        \begin{tabular}{lccccccccc}
        \toprule
                & \multicolumn{3}{c}{DPTB} & \multicolumn{3}{c}{Tiger} & \multicolumn{3}{c}{Negra} \\
                \cmidrule(lr){2-4} \cmidrule(lr){5-7} \cmidrule(lr){8-10}
                & F1 & Disc.\ F1 & POS     & F1 & Disc.\ F1 & POS      & F1 & Disc.\ F1 & POS  \\
        \midrule
        static     & 91.1 & 68.2 & 97.2    & 87.4 & 61.7 &  98.3   & 83.6 & 51.3 & 97.9   \\
        dynamic    & 91.4 & 70.9 & 97.2    & 87.6 & 62.5 &  98.4   & 84.0 & 54.0 & 98.0   \\
        \bottomrule
        \end{tabular}
    \end{center}
\caption{Results on development corpora. F1 is the Fscore on all constituents, Disc.\ F1 is an Fscore
computed only on discontinuous constituents, POS is the accuracy on part-of-speech tags.
Detailed results (including precision and recall) are given in Table~\ref{tab:details} of Appendix~\ref{sec:appendix}.}
\label{tab:dev}
\end{table*}

\begin{table*}
    \resizebox{\textwidth}{!}{
    \begin{tabular}{l cc cc cc}
        \toprule
        &\multicolumn{2}{c}{English (DPTB)} & \multicolumn{2}{c}{German (Tiger)} & \multicolumn{2}{c}{German (Negra)}  \\
        \cmidrule(lr){2-3} \cmidrule(lr){4-5} \cmidrule(lr){6-7}
        Model  & F & Disc. F & F & Disc. F & F & Disc. F \\
        \midrule
        \multicolumn{7}{c}{Predicted POS tags or own tagging} \\
        \midrule
        This work, dynamic oracle   & \bf{90.9} & 67.3 & 82.5 & \bf{55.9}  & \bf{83.2} & \bf{56.3} \\
        \midrule
        \newcite{1902.08912},$^*$ \textsc{gap}, bi-LSTM                                         & \bf{91.0} & \bf{71.3} & \bf{82.7} & \bf{55.9} & \bf{83.2} & 54.6 \\
        \newcite{stanojevic-garridoalhama:2017:EMNLP2017},$^{*}$ \textsc{swap}, stack/tree-LSTM & & & 77.0 & & & \\
        \newcite{coavoux-crabbe:2017:EACLlong}, \textsc{sr-gap}, perceptron                & &  & 79.3 & & &  \\
        \newcite{versley:2016:DiscoNLP}, pseudo-projective, chart-based                   & &  & 79.5 & & & \\
        \newcite{corro-leroux-lacroix:2017:EMNLP2017},$^{*}$ bi-LSTM, Maximum Spanning Arborescence       & 89.2    & & & & \\
        \newcite{vancranenburgh2016disc}, DOP, $\leq 40$                         & 87.0  & & & & 74.8\\
        \newcite{fernandezgonzalez-martins:2015:ACL-IJCNLP}, dependency-based             & &  & 77.3 & & & \\
        \newcite{gebhardt:2018:C18-1}, LCFRS with latent annotations   & &  & 75.1 & & & \\
        \midrule
        \multicolumn{7}{c}{Gold POS tags} \\
        \midrule
        \newcite{stanojevic-garridoalhama:2017:EMNLP2017},$^{*}$ \textsc{swap}, stack/tree-LSTM   & &  & 81.6 & & 82.9 \\
        \newcite{coavoux-crabbe:2017:EACLlong}, \textsc{sr-gap}, perceptron                       & &  & 81.6 & 49.2 & 82.2 & 50.0 \\
        \newcite{maier:2015:ACL-IJCNLP}, \textsc{swap}, perceptron   &   & & 74.7 & 18.8 &  77.0 & 19.8 \\
        \newcite{corro-leroux-lacroix:2017:EMNLP2017},$^{*}$ bi-LSTM, Maximum Spanning Arborescence  &  90.1 & & 81.6 & & \\
        \newcite{evang-kallmeyer:2011:IWPT}, PLCFRS, $< 25$ & 79$^\dagger$ & & & & \\
        \bottomrule
    \end{tabular}
    }
    \caption{Discontinuous parsing results on the test sets.\\
    $^*$Neural scoring system.
    $^\dagger$Does not discount root symbols and punctuation.}
    \label{unlex:tab:parse-test}
\end{table*}

\subsection{Objective Function}
\label{sec:objective}

In the static oracle setting, for a single sentence~$x_1^n$,
we optimize the sum
of the log-likelihood of gold POS-tags $t_1^n$
and the log-likelihood of gold parsing actions $a_1^n$:
\begin{align*}
    \mathcal{L} &= \mathcal{L}_t + \mathcal{L}_p, \\
    \mathcal{L}_{t} &= - \sum_{i=1}^n      \log P(t_i| x_1^n), \\
    \mathcal{L}_{p} &= - \sum_{i=1}^{4n-2} \log P(a_i|a_{1}^{i-1}, x_1^n).
\end{align*}
We optimize this objective by alternating a stochastic
step for the tagging objective and a stochastic step for the parsing objective,
as is standard in multitask learning \cite{Caruana:1997:ML:262868.262872}.

In the dynamic oracle setting, instead of optimizing the likelihood
of the gold actions (assuming all previous actions were gold),
we optimize the likelihood of the best actions, as computed by the dynamic
oracle, from a configuration sampled from the space of all possible configurations.
In practice, before each epoch, we sample each sentence from
the training corpus with probability $p$ and we use the current
(non-averaged) parameters to parse the sentence and generate a sequence of configurations.
Instead of selecting the highest-scoring action at each parsing step,
as in a normal inference step,
we sample an action using the softmax distribution computed by the parser,
as done by \citet{D16-1211}.
Then, we use the dynamic oracle to calculate the best action from each of these
configurations.
In case there are several best actions, we deterministically
choose a single action by favoring a \textsc{combine} over a \textsc{shift}
(to bias the model towards a small memory), and to \textsc{combine}
with the item with the highest right-index (to avoid spurious discontinuity in partial constituents).
We train the parser on these sequences of potentially non-gold configuration-action pairs.

\section{Experiments}

We carried out experiments to assess the adequacy of our system and
the effect of training with the dynamic oracle.
We present the three discontinuous constituency treebanks that we used (Section~\ref{sec:data}),
our experimental protocol (Section~\ref{sec:protocol}),
then we discuss the results (Section~\ref{sec:results})
and the efficiency of the parser (Section~\ref{sec:efficiency}).

\subsection{Datasets}
\label{sec:data}

We perform experiments on three discontinuous constituency corpora.
The discontinuous Penn Treebank was introduced by \newcite{evang-kallmeyer:2011:IWPT}
who converted the long distance dependencies encoded by indexed traces
in the original Penn treebank \cite{J93-2004}
to discontinuous constituents.
We used the standard split (sections~2-21 for training, 22~for development and 23~for test).
The Tiger corpus \cite{Brants2004} and the Negra corpus \cite{skut-EtAl:1997:ANLP}
are both German treebanks natively annotated with discontinuous constituents.
We used the SPMRL split for the Tiger corpus \cite{seddah-EtAl:2013:SPMRL},
and the split of \citet{dubey-keller:2003:ACL} for the Negra corpus.

\subsection{Implementation and Protocol}
\label{sec:protocol}

We implemented our parser in Python using the Pytorch library \cite{paszke2017automatic}.
We trained each model with the ASGD algorithm \cite{doi:10.1137/0330046}
for 100~epochs.
Training a single model takes approximately a week with a GPU.
We evaluate a model every 4~epochs on the validation set
and select the best performing model according
to the validation F-score.
We refer the reader to Table~\ref{tab:hyperparameters} of Appendix~\ref{sec:hyperparameters} for
the full list of hyperparameters.

We evaluate models with the dedicated module of
\texttt{discodop}\footnote{\url{https://github.com/andreasvc/disco-dop}}
\cite{vancranenburgh2016disc}.
We use the standard evaluation parameters (\texttt{proper.prm}),
that ignore punctuations and root symbols.
We report two evaluation metrics: a standard Fscore (F) and an Fscore computed
only on discontinuous constituents (Disc.~F), which provides a more qualitative
evaluation of the ability of the parser to recover long distance dependencies.

\subsection{Results}
\label{sec:results}

\paragraph{Effect of Dynamic Oracle}
We present parsing results on the development sets of each corpus in Table~\ref{tab:dev}.
The effect of the oracle is in line with other published results
in projective constituency parsing \cite{coavoux-crabbe:2016:P16-1,cross-huang:2016:EMNLP2016}
and dependency parsing \cite{goldberg-nivre:2012:PAPERS,gomezrodriguez-sartorio-satta:2014:EMNLP2014}:
the dynamic oracle improves the generalization capability of the parser.

\paragraph{External comparisons}
In Table~\ref{unlex:tab:parse-test}, we compare our parser to other transition-based 
parsers \cite{maier:2015:ACL-IJCNLP,coavoux-crabbe:2017:EACLlong,stanojevic-garridoalhama:2017:EMNLP2017,1902.08912},
the pseudo-projective parser of \newcite{versley:2016:DiscoNLP},
grammar-based chart parsers \cite{evang-kallmeyer:2011:IWPT,vancranenburgh2016disc,gebhardt:2018:C18-1}
and parsers based on dependency parsing \cite{fernandezgonzalez-martins:2015:ACL-IJCNLP,corro-leroux-lacroix:2017:EMNLP2017}.
Note that some of them only report results in a gold POS tag setting
(the parser has access to gold POS tags and use them as features),
a setting that is much easier than ours.

Our parser matches the state of the art of \newcite{1902.08912}.
This promising result shows that it is feasible to design accurate
transition systems without an ordered memory.

\subsection{Efficiency}
\label{sec:efficiency}

Our transition system derives a tree for a sentence of $n$ words 
in exactly $4n-2$ transitions.
Indeed, there must be $n$ \textsc{shift} actions,
and $n-1$ \textsc{combine} actions.
Each of these $2n-1$ transitions must be followed
by a single labelling action.

The statistical model responsible for choosing
which action to perform at each parsing step
needs to score $|S| + 1$ actions
for a structural step and $|N|+1$ actions
for a labelling step (where $N$ is the set of possible
nonterminals).
Since in the worst case, $|S|$ contains $n-1$ singletons,
the parser has an $\mathcal{O}(n (|N| + n))$ time complexity.

\begin{figure}[t!]
    \begin{center}
    \resizebox{\columnwidth}{!}{
    \includegraphics{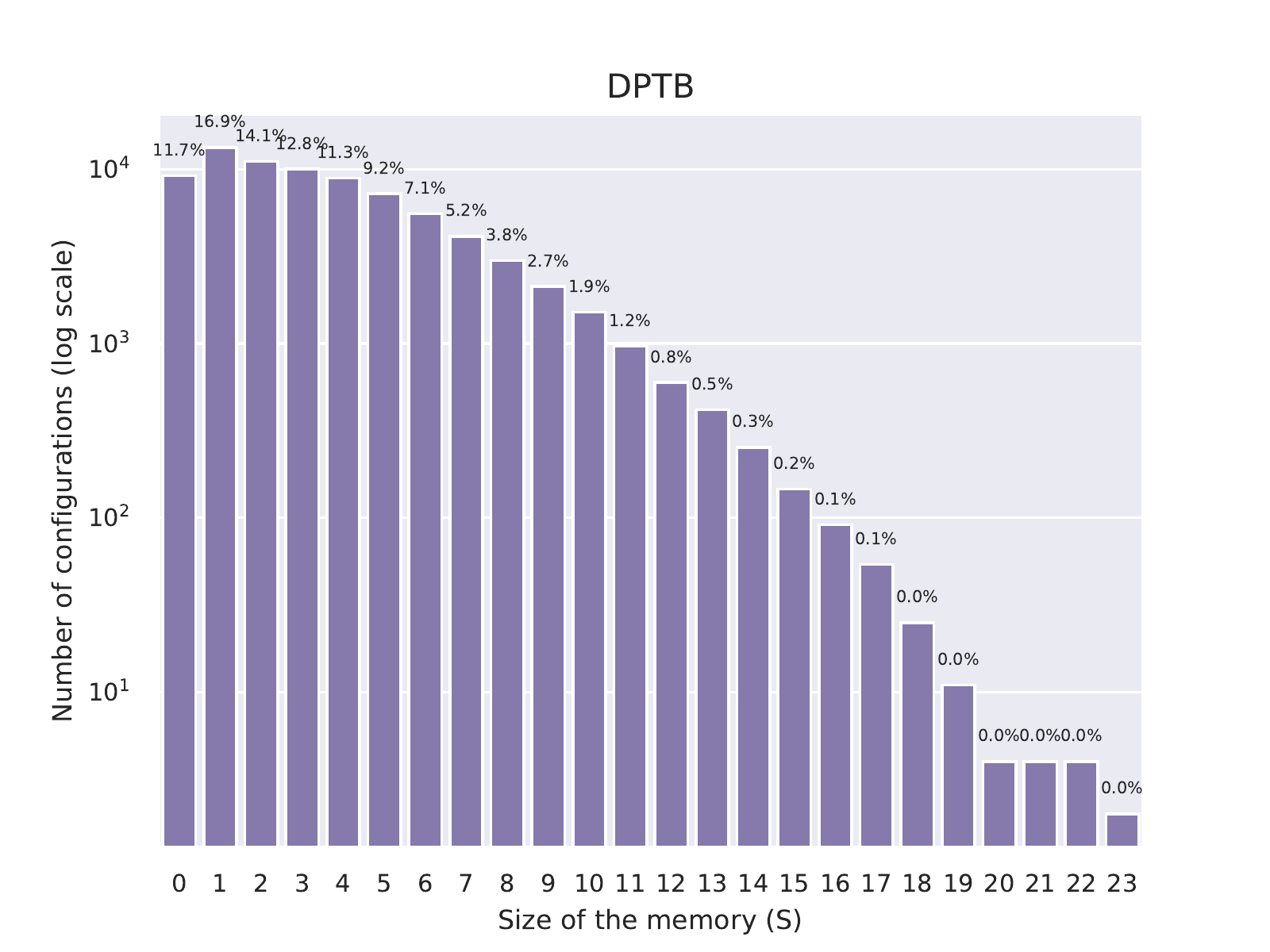}}
    \resizebox{\columnwidth}{!}{
    \includegraphics{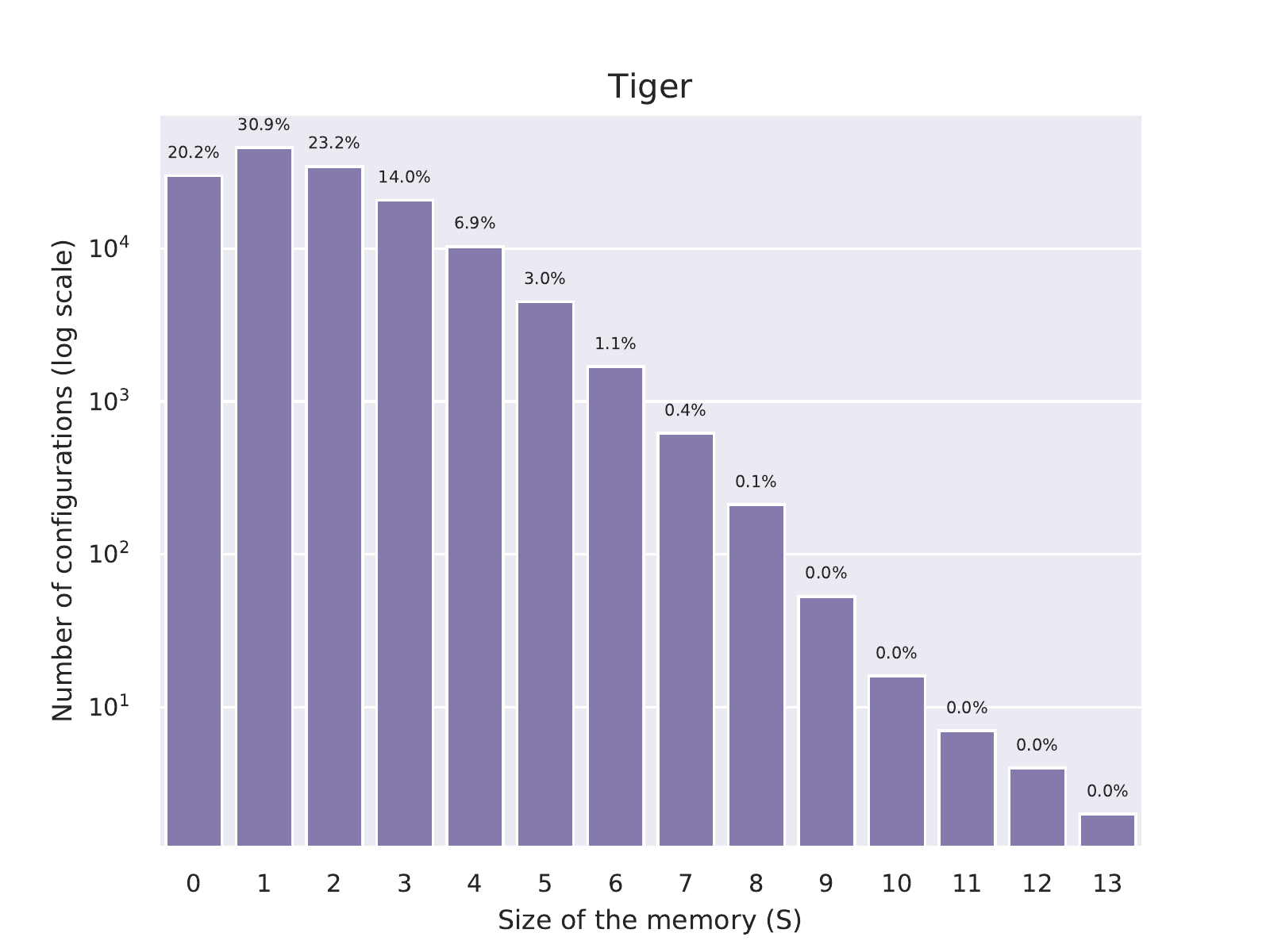}}
    \resizebox{\columnwidth}{!}{
    \includegraphics{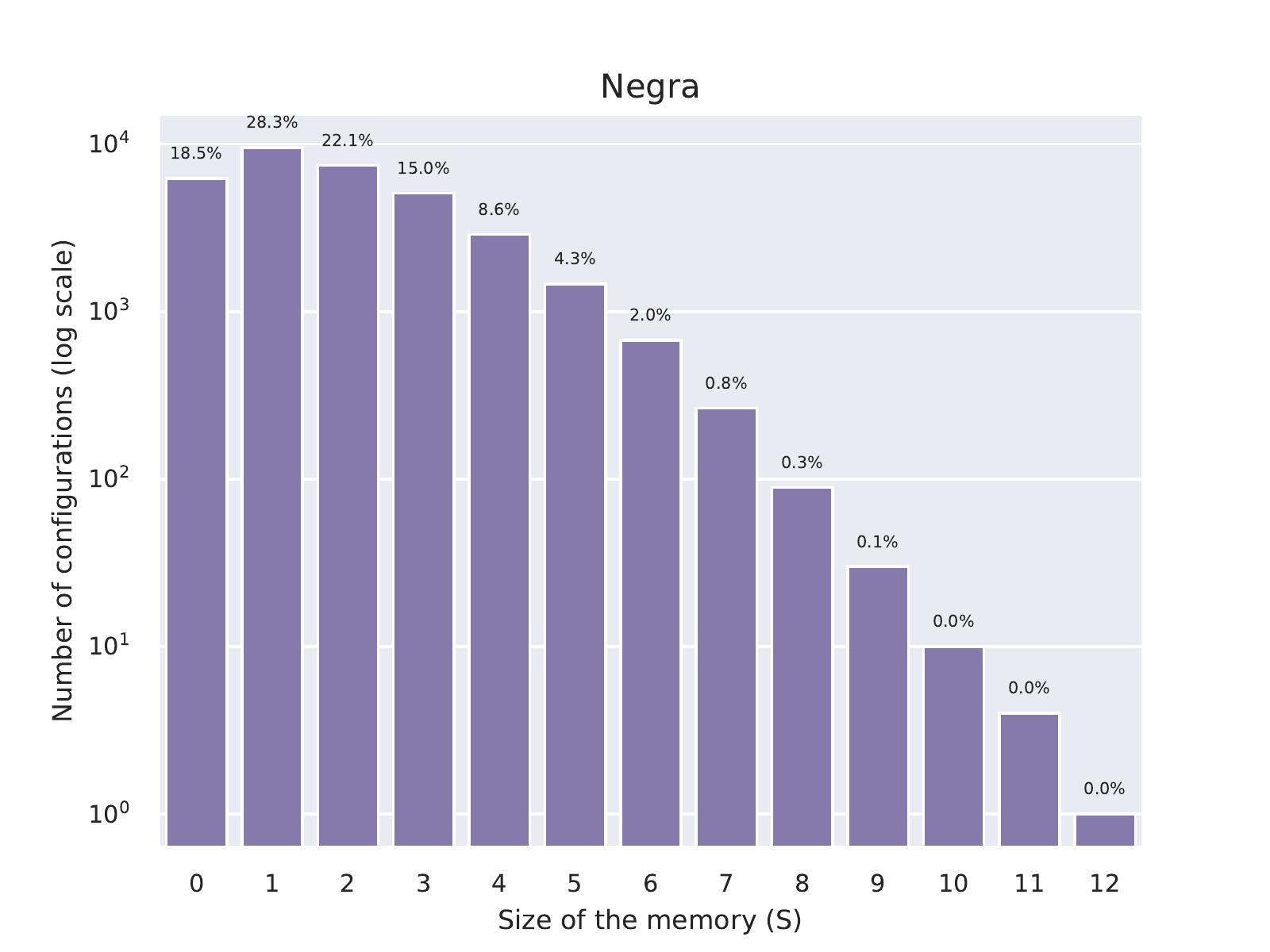}}
    \end{center}
    \caption{Distribution of the size of the memory of the parser $S$ across configurations
    derived when parsing the development set of the three corpora.
    In practice, we observe that the memory remains small.}
    \label{fig:memory}
\end{figure}

In practice, the memory of the parser $S$ remains relatively small on average.
We report in Figure~\ref{fig:memory} the distribution of the size of $S$ across configurations
when parsing the development sets of three corpora.
For the German treebanks, the memory contains 7 or fewer elements for more than 99 percents of configurations.
For the Penn treebank, the memory is slighlty larger, with 98 percents of configuration with 11 or fewer items.

We report empirical runtimes in Table~\ref{tab:parsing-times} of Appendix~\ref{sec:appendix}.
Our parser compares decently with other transition-based parsers,
despite being written in Python.

\section{Related Work}

Existing transition systems for discontinuous constituency
parsing rely on three main strategies for constructing discontinuous
constituents: a swap-based strategy, a split-stack strategy, and
the use of non-local transitions.

\paragraph{Swap-based systems}
Swap-based transition systems are based on the idea that
any discontinuous constituency tree can be transformed into
a projective tree by reordering terminals.
They reorder terminals by swapping them with a dedicated action
(\textsc{swap}), commonly used in dependency parsing \cite{nivre:2009:ACLIJCNLP}.
The first proposals in transition-based discontinuous constituency
parsing used the \textsc{swap} action on top of an easy-first parser
\cite{versley:2014:SPMRL-SANCL,DBLP:journals/corr/Versley14}.
Subsequent proposals relied on a shift-reduce system 
\cite{maier:2015:ACL-IJCNLP,maier-lichte:2016:DiscoNLP}
or a shift-promote-adjoin system \cite{stanojevic-garridoalhama:2017:EMNLP2017}.

The main limitation of swap-based system is that they tend to require a large
number of transitions to derive certain trees.
The choice of an oracle that minimizes derivation lengths has a substantially
positive effect on parsing \cite{maier-lichte:2016:DiscoNLP,stanojevic-garridoalhama:2017:EMNLP2017}.

\paragraph{Split-stack systems}
The second parsing strategy constructs discontinuous
constituents by allowing the parser to reduce pairs
of items that are not adjacent in the stack.
In practice, \newcite{coavoux-crabbe:2017:EACLlong} split
the usual stack of shift-reduce parsers
into two data structures (a stack and a double-ended queue),
in order to give the parser access to two focus items:
the respective tops of the stack and the dequeue,
that may or may not be adjacent.
A dedicated action, \textsc{gap}, pushes the top of the stack
onto the bottom of the queue to make the next item in the stack
available for a reduction.

The split stack associated with the \textsc{gap} action
can be interpreted as a linear-access memory:
it is possible to access the $i^{\text{th}}$ element
in the stack, but it requires $i$ operations.

\paragraph{Non-local transitions}

Non-local transitions generalize standard parsing actions
to non-adjacent elements in the parsing configurations.
\citet{maier-lichte:2016:DiscoNLP} introduced a non-local
transition \textsc{SkipShift}-$i$ which applies \textsc{shift}
to the $i^{\text{th}}$ element in the buffer.
Non-local transitions are also widely used in non-projective dependency parsing
\cite{W06-2922,P17-2018,N18-2109}.

The key difference between these systems and ours is that we use an unordered memory.
As a result, the semantics of the \textsc{combine}-$s$ action
we introduce in Section~\ref{sec:system} is independent
from a specific position in the stack or the buffer.
A system with an action such as \textsc{SkipShift}-$i$ needs
to learn parameters with every possible $i$, and will only learn
parameters with the \textsc{SkipShift}-$i$ actions that are required
to derive the training set.
In contrast, we use the same parameters to score each
possible \textsc{combine}-$s$ action.

\section{Conclusion}

We have presented a novel transition system that dispenses with
the use of a stack, i.e.\ a memory with linear sequential access.
Instead, the memory of the parser is represented by an unordered
data structure with random-access: a set.
We have designed a dynamic oracle for the resulting system
and shown their empirical potential with state-of-the-art results
on discontinuous constituency parsing of one English and two German treebanks.
Finally, we plan to adapt our system to non-projective dependency parsing
and semantic graph parsing.

\section*{Acknowledgments}
We thank Caio Corro, Giorgio Satta, Marco Damonte, as well as NAACL anonymous reviewers for feedback and suggestions.
We gratefully acknowledge the support of Huawei Technologies.

%% file: appendix.tex
\pagebreak
\clearpage

\appendix
\begin{table*}
\resizebox{\textwidth}{!}{
\begin{tabular}{ll}
\toprule
\multicolumn{2}{c}{Architecture hyperparameters} \\
\midrule
Dimension of word embeddings &  32 \\
Dimension of character embeddings &  100 \\
Dimension of character bi-LSTM state & 50 for each direction \\
Dimension of sentence-level bi-LSTM & 200 for each direction \\
Dimension of hidden layers for the action scorer & 200 \\
Activation functions & $\tanh$ for all hidden layers \\
\midrule
\multicolumn{2}{c}{Optimization hyperparameters} \\
\midrule
Initial learning rate   & $l_{0} = 0.01$ \\
Learning rate decay     & $l_t = \dfrac{l_{0}}{1 + t \cdot 10^{-7}}$ for step number $t$\\
Dropout for tagger input & 0.5 \\
Dropout for parser input & 0.2 \\
Training epochs         & 100 \\
Batch size              & 1 sentence \\
Optimization algorithm & Averaged SGD \cite{doi:10.1137/0330046,bottou-2010} \\
Word and character embedding initialization &  $\mathcal{U}([-0.1, 0.1])$ \\
Other parameters initialization (including LSTMs) & Xavier \cite{Glorot10understandingthe} \\
Gradient clipping (norm) & 100 \\
Dynamic oracle $p$       & 0.15 \\
\bottomrule
\end{tabular}}
\caption{Hyperparameters of the model.}
\label{tab:hyperparameters}
\end{table*}

\begin{table*}[t]
    \begin{center}
    \resizebox{0.9\textwidth}{!}{
    \begin{tabular}{llrrrr}
        \toprule
        Parser                                  & Setting      & \multicolumn{2}{c}{Tiger} & \multicolumn{2}{c}{DPTB} \\
        \cmidrule(lr){3-4}   \cmidrule(lr){5-6}
                                                &              & tok/s & sent/s         & tok/s &  sent/s \\
        \midrule
        This work                               & Python, neural, greedy, CPU    & 978  & 64  &  910 & 38 \\
        \midrule
        MTG \cite{1902.08912}                   & C++, neural, greedy, CPU       & 1934 & 126 & 1887 & 80 \\
        MTG \cite{coavoux-crabbe:2017:EACLlong} & C++, perceptron, beam=4, CPU   & 4700 & 260 \\
        rparse \cite{maier:2015:ACL-IJCNLP}     & Java, perceptron, beam=8, CPU  & & 80 \\
        rparse \cite{maier:2015:ACL-IJCNLP}     & Java, perceptron, beam=1, CPU  & & 640 \\
        \newcite{corro-leroux-lacroix:2017:EMNLP2017} & C++, neural, CPU         & & & & $\approx 7.3$ \\
        \bottomrule
    \end{tabular}}
    \end{center}
    \caption{Parsing times on development sets in tokens per second (tok/s) and sentences per second (sent/s).
    The parsing times are presented as reported by authors, they are not comparable across parsers
    (since the experiments were run on different hardware).
    Our parser is run on a single core of an Intel i7 CPU.}
    \label{tab:parsing-times}
\end{table*}

\section{Oracle Correctness}
\label{sec:correctness}
The oracle $\mathsf{o}$ leads
to the reachable tree with the highest F-score with respect
to the gold tree.
The F-score of a predicted tree $\hat t$ (represented as a set of instantiated constituents)
with respect to a gold tree $t^*$ is defined as:
\begin{align*}
    \mathsf{precision}(\hat t, t^*) &= p = \frac{|\hat t \cap t^*|}{|\hat t|}, \\
    \mathsf{recall}(\hat t, t^*) &= r = \frac{|\hat t \cap t^*|}{|t^*|}, \\
    F_1(\hat t, t^*) &= \frac{2pr}{p+r}.
\end{align*}
By definition, $\mathsf{o_{odd}}$ is optimal for precision
because it constructs a constituent only if it is gold,
and optimal for recall because it will construct a gold
constituent if it is possible to do so.

Moreover, $\mathsf{o_{even}}$ is optimal for recall because
any gold constituent reachable from $c$ will still be reachable
after any transition in $\mathsf{o_{even}}(c, t^*)$.
Assuming a configuration $c = (S, s_f, i, C)$
and $\mathsf{next}(c, t^*)=s_g$, we consider
separately the \textsc{shift} case and the \textsc{combine}-$s$ case:
\begin{itemize}[noitemsep]
    \item \textsc{shift} case ($\max(s_g) > \max(s_f)$): constituents $(X, s)$
    reachable from $c$ and not reachable from \textsc{shift}($c$)
    satisfy $\max(s)=i$. 
    If a gold constituent satisfies this property, we have $s \preceq s_g$, which
    contradicts the assumption that $s_g = \mathsf{next}(c, t^*)$ (see definition of oracle in Section~\ref{sec:dyno}).
    \item \textsc{combine}-$s$ case:
    Let $(X, s')$ be a reachable gold constituent.
    Since it is compatible with $s_g$, there are three possible cases:
    \begin{itemize}[noitemsep]
        \item if $(X, s')$ is an ascendant of $s_g$,
        then $(s \cup s_f) \subseteq s_g \subset s'$,
        therefore $(X, s')$ is still reachable from \textsc{combine}-$s$($c$).
        \item if $(X, s')$ is a descendant of $s_g$ then $s' \preceq s_g$,
        which contradicts the definition of $s_g$.
        \item if $s'$ and $s_g$ are completely disjoint, 
        we have $s' \cap s = s' \cap s_f = \emptyset$,
        therefore $s' \cap (s \cup s_f) = \emptyset$,
        and $s'$ is still reachable from \textsc{combine}-$s$($c$).
    \end{itemize}
\end{itemize}
Finally, since $\mathsf{o_{even}}$ does not construct
new constituents (it is the role of labelling actions),
it is optimal for precision.

\section{Hyperparameters}
\label{sec:hyperparameters}

The list of hyperparameters is presented in Table~\ref{tab:hyperparameters}.

\begin{itemize}[noitemsep]
    \item We use learning rate warm-up (linear increase from 0 to $t_{1000}$ during the first 1000 steps).
    \item Before the $t^{th}$ update, we add Gaussian noise to the gradient of every parameter with mean 0 and variance $\dfrac{0.01}{(1 + t)^{0.55}}$ \cite{DBLP:journals/corr/NeelakantanVLSK15}.
    \item All experiments use greedy search decoding (we did not experiment with beam search).
    \item Before each training step, we replace a word embedding by an `UNK' pseudo-word embedding
    with probability $0.3$. We only do this replacement for the least frequent
    word-types ($\frac{2}{3}$ least frequent word-types). The `UNK' embedding
    is then used to represent unknown words.
    \item We apply dropout at the input of the tagger and the input of action scorers:
    each single prediction has its own dropout mask.
\end{itemize}

\section{Detailed Results}
\label{sec:appendix}

\begin{table*}
\centering
\begin{tabular}{ll|lll|lll|l}
    \toprule
        && \multicolumn{3}{c|}{All const.} & \multicolumn{3}{c|}{Disc.\ const.} & POS \\
    Development sets    && F & P & R
                        & F & P & R & Acc.\\
    \midrule
    English (DPTB) & static  & 91.1 & 91.1 & 91.2 & 68.2 & 75.3 & 62.3 &  97.2 \\
                   & dynamic & 91.4 & 91.5 & 91.3 & 70.9 & 76.1 & 66.4 & 97.2  \\
    \midrule
    German (Tiger) & static  & 87.4 & 87.8 & 87.0 & 61.7 & 64.4 & 59.2 &  98.3 \\
                   & dynamic & 87.6 & 88.2 & 87.0 & 62.5 & 68.6 & 57.3 & 98.4  \\
    \midrule
    German (Negra) & static  & 83.6 & 83.8 & 83.4 & 51.3 & 53.3 & 49.5 &  97.9 \\
                   & dynamic & 84.0 & 84.7 & 83.4 & 54.0 & 58.1 & 50.5 & 98.0  \\
    \midrule
    Test sets    && F & P & R
                        & F & P & R & Acc.\\
    \midrule
    English (DPTB) & dynamic & 90.9 & 91.3 & 90.6 & 67.3 & 73.3 & 62.1 &  97.6 \\
    German (Tiger) & dynamic & 82.5 & 83.5 & 81.5 & 55.9 & 62.4 & 50.6 &  98.0 \\
    German (Negra) & dynamic & 83.2 & 83.8 & 82.6 & 56.3 & 64.9 & 49.8 &  98.0 \\
    \bottomrule
\end{tabular}
\caption{Detailed results.
Overall, the positive effect of the dynamic oracle on Fscore
is explained by its effect on precision.}
\label{tab:details}
\end{table*}

%% file: naaclhlt2019.bbl
\begin{thebibliography}{47}
\expandafter\ifx\csname natexlab\endcsname\relax\def\natexlab#1{#1}\fi

\bibitem[{Attardi(2006)}]{W06-2922}
Giuseppe Attardi. 2006.
\newblock \href {http://www.aclweb.org/anthology/W06-2922} {Experiments with a
  multilanguage non-projective dependency parser}.
\newblock In \emph{Proceedings of the Tenth Conference on Computational Natural
  Language Learning (CoNLL-X)}, pages 166--170. Association for Computational
  Linguistics.

\bibitem[{Ballesteros et~al.(2016)Ballesteros, Goldberg, Dyer, and
  Smith}]{D16-1211}
Miguel Ballesteros, Yoav Goldberg, Chris Dyer, and Noah~A. Smith. 2016.
\newblock \href {https://doi.org/10.18653/v1/D16-1211} {Training with
  exploration improves a greedy stack lstm parser}.
\newblock In \emph{Proceedings of the 2016 Conference on Empirical Methods in
  Natural Language Processing}, pages 2005--2010, Austin, Texas. Association
  for Computational Linguistics.

\bibitem[{Bottou(2010)}]{bottou-2010}
L\'{e}on Bottou. 2010.
\newblock \href {http://leon.bottou.org/papers/bottou-2010} {Large-scale
  machine learning with stochastic gradient descent}.
\newblock In \emph{Proceedings of the 19th International Conference on
  Computational Statistics (COMPSTAT'2010)}, pages 177--187, Paris, France.
  Springer.

\bibitem[{Brants et~al.(2004)Brants, Dipper, Eisenberg, Hansen-Schirra,
  K{\"o}nig, Lezius, Rohrer, Smith, and Uszkoreit}]{Brants2004}
Sabine Brants, Stefanie Dipper, Peter Eisenberg, Silvia Hansen-Schirra, Esther
  K{\"o}nig, Wolfgang Lezius, Christian Rohrer, George Smith, and Hans
  Uszkoreit. 2004.
\newblock \href {https://doi.org/10.1007/s11168-004-7431-3} {Tiger: Linguistic
  interpretation of a german corpus}.
\newblock \emph{Research on Language and Computation}, 2(4):597--620.

\bibitem[{Caruana(1997)}]{Caruana:1997:ML:262868.262872}
Rich Caruana. 1997.
\newblock \href {https://doi.org/10.1023/A:1007379606734} {Multitask learning}.
\newblock \emph{Machine Learning}, 28(1):41--75.

\bibitem[{Coavoux and Crabb\'{e}(2016)}]{coavoux-crabbe:2016:P16-1}
Maximin Coavoux and Benoit Crabb\'{e}. 2016.
\newblock \href {http://www.aclweb.org/anthology/P16-1017} {Neural greedy
  constituent parsing with dynamic oracles}.
\newblock In \emph{Proceedings of the 54th Annual Meeting of the Association
  for Computational Linguistics (Volume 1: Long Papers)}, pages 172--182,
  Berlin, Germany. Association for Computational Linguistics.

\bibitem[{Coavoux and
  Crabb\'{e}(2017{\natexlab{a}})}]{coavoux-crabbe:2017:EACLlong}
Maximin Coavoux and Benoit Crabb\'{e}. 2017{\natexlab{a}}.
\newblock \href {http://www.aclweb.org/anthology/E17-1118} {Incremental
  discontinuous phrase structure parsing with the gap transition}.
\newblock In \emph{Proceedings of the 15th Conference of the European Chapter
  of the Association for Computational Linguistics: Volume 1, Long Papers},
  pages 1259--1270, Valencia, Spain. Association for Computational Linguistics.

\bibitem[{Coavoux and
  Crabb\'{e}(2017{\natexlab{b}})}]{coavoux-crabbe:2017:EACLshort}
Maximin Coavoux and Benoit Crabb\'{e}. 2017{\natexlab{b}}.
\newblock \href {http://www.aclweb.org/anthology/E17-2053} {Multilingual
  lexicalized constituency parsing with word-level auxiliary tasks}.
\newblock In \emph{Proceedings of the 15th Conference of the European Chapter
  of the Association for Computational Linguistics: Volume 2, Short Papers},
  pages 331--336, Valencia, Spain. Association for Computational Linguistics.

\bibitem[{Coavoux et~al.(2019)Coavoux, Crabb\'{e}, and Cohen}]{1902.08912}
Maximin Coavoux, Beno\^{i}t Crabb\'{e}, and Shay~B. Cohen. 2019.
\newblock \href {http://arxiv.org/abs/arXiv:1902.08912v1} {Unlexicalized
  transition-based discontinuous constituency parsing}.
\newblock \emph{CoRR}, abs/1902.08912v1.

\bibitem[{Cohen et~al.(2012)Cohen, G{\'{o}}mez{-}Rodr{\'{\i}}guez, and
  Satta}]{DBLP:journals/corr/abs-1206-6735}
Shay~B. Cohen, Carlos G{\'{o}}mez{-}Rodr{\'{\i}}guez, and Giorgio Satta. 2012.
\newblock \href {http://arxiv.org/abs/1206.6735} {Elimination of spurious
  ambiguity in transition-based dependency parsing}.
\newblock \emph{CoRR}, abs/1206.6735.

\bibitem[{Corro et~al.(2017)Corro, Le~Roux, and
  Lacroix}]{corro-leroux-lacroix:2017:EMNLP2017}
Caio Corro, Joseph Le~Roux, and Mathieu Lacroix. 2017.
\newblock \href {https://www.aclweb.org/anthology/D17-1172} {Efficient
  discontinuous phrase-structure parsing via the generalized maximum spanning
  arborescence}.
\newblock In \emph{Proceedings of the 2017 Conference on Empirical Methods in
  Natural Language Processing}, pages 1645--1655, Copenhagen, Denmark.
  Association for Computational Linguistics.

\bibitem[{Crabb\'{e}(2015)}]{crabbe:2015:EMNLP}
Benoit Crabb\'{e}. 2015.
\newblock \href {http://aclweb.org/anthology/D15-1212} {Multilingual
  discriminative lexicalized phrase structure parsing}.
\newblock In \emph{Proceedings of the 2015 Conference on Empirical Methods in
  Natural Language Processing}, pages 1847--1856, Lisbon, Portugal. Association
  for Computational Linguistics.

\bibitem[{van Cranenburgh et~al.(2016)van Cranenburgh, Scha, and
  Bod}]{vancranenburgh2016disc}
Andreas van Cranenburgh, Remko Scha, and Rens Bod. 2016.
\newblock \href {http://dx.doi.org/10.15398/jlm.v4i1.100} {Data-oriented
  parsing with discontinuous constituents and function tags}.
\newblock \emph{Journal of Language Modelling}, 4(1):57--111.

\bibitem[{Cross and Huang(2016{\natexlab{a}})}]{cross-huang:2016:P16-2}
James Cross and Liang Huang. 2016{\natexlab{a}}.
\newblock \href {http://anthology.aclweb.org/P16-2006} {Incremental parsing
  with minimal features using bi-directional {LSTM}}.
\newblock In \emph{Proceedings of the 54th Annual Meeting of the Association
  for Computational Linguistics (Volume 2: Short Papers)}, pages 32--37,
  Berlin, Germany. Association for Computational Linguistics.

\bibitem[{Cross and Huang(2016{\natexlab{b}})}]{cross-huang:2016:EMNLP2016}
James Cross and Liang Huang. 2016{\natexlab{b}}.
\newblock \href {https://aclweb.org/anthology/D16-1001} {Span-based
  constituency parsing with a structure-label system and provably optimal
  dynamic oracles}.
\newblock In \emph{Proceedings of the 2016 Conference on Empirical Methods in
  Natural Language Processing}, pages 1--11, Austin, Texas. Association for
  Computational Linguistics.

\bibitem[{Dubey and Keller(2003)}]{dubey-keller:2003:ACL}
Amit Dubey and Frank Keller. 2003.
\newblock \href {https://doi.org/10.3115/1075096.1075109} {Probabilistic
  parsing for german using sister-head dependencies}.
\newblock In \emph{Proceedings of the 41st Annual Meeting of the Association
  for Computational Linguistics}, pages 96--103, Sapporo, Japan. Association
  for Computational Linguistics.

\bibitem[{Durrett and Klein(2015)}]{durrett-klein:2015:ACL-IJCNLP}
Greg Durrett and Dan Klein. 2015.
\newblock \href {http://www.aclweb.org/anthology/P15-1030} {Neural {CRF}
  parsing}.
\newblock In \emph{Proceedings of the 53rd Annual Meeting of the Association
  for Computational Linguistics and the 7th International Joint Conference on
  Natural Language Processing (Volume 1: Long Papers)}, pages 302--312,
  Beijing, China. Association for Computational Linguistics.

\bibitem[{Evang and Kallmeyer(2011)}]{evang-kallmeyer:2011:IWPT}
Kilian Evang and Laura Kallmeyer. 2011.
\newblock \href {http://www.aclweb.org/anthology/W11-2913} {{PLCFRS} parsing of
  english discontinuous constituents}.
\newblock In \emph{Proceedings of the 12th International Conference on Parsing
  Technologies}, pages 104--116, Dublin, Ireland. Association for Computational
  Linguistics.

\bibitem[{Fern\'{a}ndez-Gonz\'{a}lez and
  G\'{o}mez-Rodr\'{i}guez(2018)}]{N18-2109}
Daniel Fern\'{a}ndez-Gonz\'{a}lez and Carlos G\'{o}mez-Rodr\'{i}guez. 2018.
\newblock \href {https://doi.org/10.18653/v1/N18-2109} {Non-projective
  dependency parsing with non-local transitions}.
\newblock In \emph{Proceedings of the 2018 Conference of the North American
  Chapter of the Association for Computational Linguistics: Human Language
  Technologies, Volume 2 (Short Papers)}, pages 693--700. Association for
  Computational Linguistics.

\bibitem[{Fern\'{a}ndez-Gonz\'{a}lez and
  G\'{o}mez-Rodr\'{i}­guez(2018{\natexlab{a}})}]{fernndezgonzlez-gmezrodrguez:2018:EMNLP}
Daniel Fern\'{a}ndez-Gonz\'{a}lez and Carlos G\'{o}mez-Rodr\'{i}­guez.
  2018{\natexlab{a}}.
\newblock \href {http://www.aclweb.org/anthology/D18-1161} {Dynamic oracles for
  top-down and in-order shift-reduce constituent parsing}.
\newblock In \emph{Proceedings of the 2018 Conference on Empirical Methods in
  Natural Language Processing}, pages 1303--1313, Brussels, Belgium.
  Association for Computational Linguistics.

\bibitem[{Fern\'{a}ndez-Gonz\'{a}lez and
  G\'{o}mez-Rodr\'{i}­guez(2018{\natexlab{b}})}]{DBLP:journals/corr/abs-1804-07961}
Daniel Fern\'{a}ndez-Gonz\'{a}lez and Carlos G\'{o}mez-Rodr\'{i}­guez.
  2018{\natexlab{b}}.
\newblock \href {http://arxiv.org/abs/1804.07961} {Faster shift-reduce
  constituent parsing with a non-binary, bottom-up strategy}.
\newblock \emph{CoRR}, abs/1804.07961.

\bibitem[{Fern\'{a}ndez-Gonz\'{a}lez and
  Martins(2015)}]{fernandezgonzalez-martins:2015:ACL-IJCNLP}
Daniel Fern\'{a}ndez-Gonz\'{a}lez and Andr\'{e} F.~T. Martins. 2015.
\newblock \href {http://www.aclweb.org/anthology/P15-1147} {Parsing as
  reduction}.
\newblock In \emph{Proceedings of the 53rd Annual Meeting of the Association
  for Computational Linguistics and the 7th International Joint Conference on
  Natural Language Processing (Volume 1: Long Papers)}, pages 1523--1533,
  Beijing, China. Association for Computational Linguistics.

\bibitem[{Gebhardt(2018)}]{gebhardt:2018:C18-1}
Kilian Gebhardt. 2018.
\newblock \href {http://www.aclweb.org/anthology/C18-1258} {Generic refinement
  of expressive grammar formalisms with an application to discontinuous
  constituent parsing}.
\newblock In \emph{Proceedings of the 27th International Conference on
  Computational Linguistics}, pages 3049--3063, Santa Fe, New Mexico, USA.
  Association for Computational Linguistics.

\bibitem[{Glorot and Bengio(2010)}]{Glorot10understandingthe}
Xavier Glorot and Yoshua Bengio. 2010.
\newblock Understanding the difficulty of training deep feedforward neural
  networks.
\newblock In \emph{In Proceedings of the International Conference on Artificial
  Intelligence and Statistics (AISTATS’10). Society for Artificial
  Intelligence and Statistics}.

\bibitem[{Goldberg and Nivre(2012)}]{goldberg-nivre:2012:PAPERS}
Yoav Goldberg and Joakim Nivre. 2012.
\newblock \href {http://www.aclweb.org/anthology/C12-1059} {A dynamic oracle
  for arc-eager dependency parsing}.
\newblock In \emph{Proceedings of COLING 2012}, pages 959--976, Mumbai, India.
  The COLING 2012 Organizing Committee.

\bibitem[{Goldberg and Nivre(2013)}]{Q13-1033}
Yoav Goldberg and Joakim Nivre. 2013.
\newblock \href {http://aclweb.org/anthology/Q13-1033} {Training deterministic
  parsers with non-deterministic oracles}.
\newblock \emph{Transactions of the Association for Computational Linguistics},
  1:403--414.

\bibitem[{G\'{o}mez-Rodr\'{i}guez and
  Fern\'{a}ndez-Gonz\'{a}lez(2015)}]{gomezrodriguez-fernandezgonzalez:2015:ACL-IJCNLP}
Carlos G\'{o}mez-Rodr\'{i}guez and Daniel Fern\'{a}ndez-Gonz\'{a}lez. 2015.
\newblock \href {http://www.aclweb.org/anthology/P15-2042} {An efficient
  dynamic oracle for unrestricted non-projective parsing}.
\newblock In \emph{Proceedings of the 53rd Annual Meeting of the Association
  for Computational Linguistics and the 7th International Joint Conference on
  Natural Language Processing (Volume 2: Short Papers)}, pages 256--261,
  Beijing, China. Association for Computational Linguistics.

\bibitem[{G\'{o}mez-Rodr\'{i}guez et~al.(2014)G\'{o}mez-Rodr\'{i}guez,
  Sartorio, and Satta}]{gomezrodriguez-sartorio-satta:2014:EMNLP2014}
Carlos G\'{o}mez-Rodr\'{i}guez, Francesco Sartorio, and Giorgio Satta. 2014.
\newblock \href {http://www.aclweb.org/anthology/D14-1099} {A polynomial-time
  dynamic oracle for non-projective dependency parsing}.
\newblock In \emph{Proceedings of the 2014 Conference on Empirical Methods in
  Natural Language Processing (EMNLP)}, pages 917--927, Doha, Qatar.
  Association for Computational Linguistics.

\bibitem[{Hall et~al.(2014)Hall, Durrett, and
  Klein}]{hall-durrett-klein:2014:P14-1}
David Hall, Greg Durrett, and Dan Klein. 2014.
\newblock \href {http://www.aclweb.org/anthology/P14-1022} {Less grammar, more
  features}.
\newblock In \emph{Proceedings of the 52nd Annual Meeting of the Association
  for Computational Linguistics (Volume 1: Long Papers)}, pages 228--237,
  Baltimore, Maryland. Association for Computational Linguistics.

\bibitem[{Kallmeyer(2010)}]{kallmeyer-book}
Laura Kallmeyer. 2010.
\newblock \emph{Parsing Beyond Context-Free Grammars}, 1st edition.
\newblock Springer Publishing Company, Incorporated.

\bibitem[{Kiperwasser and Goldberg(2016)}]{TACL885}
Eliyahu Kiperwasser and Yoav Goldberg. 2016.
\newblock \href {https://transacl.org/ojs/index.php/tacl/article/view/885}
  {Simple and accurate dependency parsing using bidirectional {LSTM} feature
  representations}.
\newblock \emph{Transactions of the Association for Computational Linguistics},
  4:313--327.

\bibitem[{Maier(2015)}]{maier:2015:ACL-IJCNLP}
Wolfgang Maier. 2015.
\newblock \href {http://www.aclweb.org/anthology/P15-1116} {Discontinuous
  incremental shift-reduce parsing}.
\newblock In \emph{Proceedings of the 53rd Annual Meeting of the Association
  for Computational Linguistics and the 7th International Joint Conference on
  Natural Language Processing (Volume 1: Long Papers)}, pages 1202--1212,
  Beijing, China. Association for Computational Linguistics.

\bibitem[{Maier and Lichte(2016)}]{maier-lichte:2016:DiscoNLP}
Wolfgang Maier and Timm Lichte. 2016.
\newblock \href {http://www.aclweb.org/anthology/W16-0906} {Discontinuous
  parsing with continuous trees}.
\newblock In \emph{Proceedings of the Workshop on Discontinuous Structures in
  Natural Language Processing}, pages 47--57, San Diego, California.
  Association for Computational Linguistics.

\bibitem[{Marcus et~al.(1993)Marcus, Santorini, and
  Ann~Marcinkiewicz}]{J93-2004}
Mitchell Marcus, Beatrice Santorini, and Mary Ann~Marcinkiewicz. 1993.
\newblock \href {http://aclweb.org/anthology/J93-2004} {Building a large
  annotated corpus of english: The penn treebank}.
\newblock \emph{Computational Linguistics, Volume 19, Number 2, June 1993,
  Special Issue on Using Large Corpora: II}.

\bibitem[{Neelakantan et~al.(2015)Neelakantan, Vilnis, Le, Sutskever, Kaiser,
  Kurach, and Martens}]{DBLP:journals/corr/NeelakantanVLSK15}
Arvind Neelakantan, Luke Vilnis, Quoc~V. Le, Ilya Sutskever, Lukasz Kaiser,
  Karol Kurach, and James Martens. 2015.
\newblock \href {http://arxiv.org/abs/1511.06807} {Adding gradient noise
  improves learning for very deep networks}.
\newblock \emph{CoRR}, abs/1511.06807.

\bibitem[{Nivre(2009)}]{nivre:2009:ACLIJCNLP}
Joakim Nivre. 2009.
\newblock \href {http://www.aclweb.org/anthology/P/P09/P09-1040}
  {Non-projective dependency parsing in expected linear time}.
\newblock In \emph{Proceedings of the Joint Conference of the 47th Annual
  Meeting of the ACL and the 4th International Joint Conference on Natural
  Language Processing of the AFNLP}, pages 351--359, Suntec, Singapore.
  Association for Computational Linguistics.

\bibitem[{Paszke et~al.(2017)Paszke, Gross, Chintala, Chanan, Yang, DeVito,
  Lin, Desmaison, Antiga, and Lerer}]{paszke2017automatic}
Adam Paszke, Sam Gross, Soumith Chintala, Gregory Chanan, Edward Yang, Zachary
  DeVito, Zeming Lin, Alban Desmaison, Luca Antiga, and Adam Lerer. 2017.
\newblock Automatic differentiation in pytorch.
\newblock In \emph{NIPS-W}.

\bibitem[{Polyak and Juditsky(1992)}]{doi:10.1137/0330046}
Boris~T. Polyak and Anatoli~B. Juditsky. 1992.
\newblock \href {https://doi.org/10.1137/0330046} {Acceleration of stochastic
  approximation by averaging}.
\newblock \emph{SIAM Journal on Control and Optimization}, 30(4):838--855.

\bibitem[{Qi and Manning(2017)}]{P17-2018}
Peng Qi and Christopher~D. Manning. 2017.
\newblock \href {https://doi.org/10.18653/v1/P17-2018} {Arc-swift: A novel
  transition system for dependency parsing}.
\newblock In \emph{Proceedings of the 55th Annual Meeting of the Association
  for Computational Linguistics (Volume 2: Short Papers)}, pages 110--117.
  Association for Computational Linguistics.

\bibitem[{Seddah et~al.(2013)Seddah, Tsarfaty, K\"{u}bler, Candito, Choi,
  Farkas, Foster, Goenaga, Gojenola~Galletebeitia, Goldberg, Green, Habash,
  Kuhlmann, Maier, Nivre, Przepi\'{o}rkowski, Roth, Seeker, Versley, Vincze,
  Woli\'{n}ski, Wr\'{o}blewska, and de~la Clergerie}]{seddah-EtAl:2013:SPMRL}
Djam\'{e} Seddah, Reut Tsarfaty, Sandra K\"{u}bler, Marie Candito, Jinho~D.
  Choi, Rich\'{a}rd Farkas, Jennifer Foster, Iakes Goenaga, Koldo
  Gojenola~Galletebeitia, Yoav Goldberg, Spence Green, Nizar Habash, Marco
  Kuhlmann, Wolfgang Maier, Joakim Nivre, Adam Przepi\'{o}rkowski, Ryan Roth,
  Wolfgang Seeker, Yannick Versley, Veronika Vincze, Marcin Woli\'{n}ski, Alina
  Wr\'{o}blewska, and Eric~Villemonte de~la Clergerie. 2013.
\newblock \href {http://www.aclweb.org/anthology/W13-4917} {Overview of the
  {SPMRL} 2013 shared task: A cross-framework evaluation of parsing
  morphologically rich languages}.
\newblock In \emph{Proceedings of the Fourth Workshop on Statistical Parsing of
  Morphologically-Rich Languages}, pages 146--182, Seattle, Washington, USA.
  Association for Computational Linguistics.

\bibitem[{Skut et~al.(1997)Skut, Krenn, Brants, and
  Uszkoreit}]{skut-EtAl:1997:ANLP}
Wojciech Skut, Brigitte Krenn, Thorsten Brants, and Hans Uszkoreit. 1997.
\newblock \href {https://doi.org/10.3115/974557.974571} {An annotation scheme
  for free word order languages}.
\newblock In \emph{Proceedings of the Fifth Conference on Applied Natural
  Language Processing}, pages 88--95, Washington, DC, USA. Association for
  Computational Linguistics.

\bibitem[{Stanojevi\'{c} and
  Garrido~Alhama(2017)}]{stanojevic-garridoalhama:2017:EMNLP2017}
Milo\v{s} Stanojevi\'{c} and Raquel Garrido~Alhama. 2017.
\newblock \href {https://www.aclweb.org/anthology/D17-1174} {Neural
  discontinuous constituency parsing}.
\newblock In \emph{Proceedings of the 2017 Conference on Empirical Methods in
  Natural Language Processing}, pages 1667--1677, Copenhagen, Denmark.
  Association for Computational Linguistics.

\bibitem[{Stern et~al.(2017)Stern, Andreas, and
  Klein}]{stern-andreas-klein:2017:Long}
Mitchell Stern, Jacob Andreas, and Dan Klein. 2017.
\newblock \href {http://aclweb.org/anthology/P17-1076} {A minimal span-based
  neural constituency parser}.
\newblock In \emph{Proceedings of the 55th Annual Meeting of the Association
  for Computational Linguistics (Volume 1: Long Papers)}, pages 818--827,
  Vancouver, Canada. Association for Computational Linguistics.

\bibitem[{Versley(2014{\natexlab{a}})}]{versley:2014:SPMRL-SANCL}
Yannick Versley. 2014{\natexlab{a}}.
\newblock \href {http://www.aclweb.org/anthology/W14-6104} {Experiments with
  easy-first nonprojective constituent parsing}.
\newblock In \emph{Proceedings of the First Joint Workshop on Statistical
  Parsing of Morphologically Rich Languages and Syntactic Analysis of
  Non-Canonical Languages}, pages 39--53, Dublin, Ireland. Dublin City
  University.

\bibitem[{Versley(2014{\natexlab{b}})}]{DBLP:journals/corr/Versley14}
Yannick Versley. 2014{\natexlab{b}}.
\newblock \href {http://arxiv.org/abs/1409.3813} {Incorporating semi-supervised
  features into discontinuous easy-first constituent parsing}.
\newblock \emph{CoRR}, abs/1409.3813.

\bibitem[{Versley(2016)}]{versley:2016:DiscoNLP}
Yannick Versley. 2016.
\newblock \href {http://www.aclweb.org/anthology/W16-0907} {Discontinuity
  re\^{}2-visited: A minimalist approach to pseudoprojective constituent
  parsing}.
\newblock In \emph{Proceedings of the Workshop on Discontinuous Structures in
  Natural Language Processing}, pages 58--69, San Diego, California.
  Association for Computational Linguistics.

\bibitem[{Vijay-Shanker et~al.(1987)Vijay-Shanker, Weir, and Joshi}]{P87-1015}
K.~Vijay-Shanker, David~J. Weir, and Aravind~K. Joshi. 1987.
\newblock \href
  {http://aclanthology.coli.uni-saarland.de/pdf/P/P87/P87-1015.pdf}
  {Characterizing structural descriptions produced by various grammatical
  formalisms}.
\newblock In \emph{25th Annual Meeting of the Association for Computational
  Linguistics}.

\end{thebibliography}
